# Detecting Activities of Daily Living in Egocentric Video to Contextualize Hand Use at Home in Outpatient Neurorehabilitation Settings

Adesh Kadambi[1,2], José Zariffa[1,2,3,4*]

[1]Institute of Biomedical Engineering, University of Toronto, Toronto, ON, Canada
[2]KITE - Toronto Rehabilitation Institute, University Health Network, Toronto, ON, Canada
[3]Rehabilitation Sciences Institute, University of Toronto, Toronto, ON, Canada
[4]Edward S. Rogers Sr. Department of Electrical and Computer Engineering, University of Toronto, Toronto, ON, Canada

## Abstract

Wearable egocentric cameras and machine learning have the potential to provide clinicians with a more nuanced understanding of patient hand use at home after stroke and spinal cord injury (SCI). However, they require detailed contextual information (i.e., activities and object interactions) to effectively interpret metrics and meaningfully guide therapy planning. We demonstrate that an object-centric approach, focusing on what objects patients interact with rather than how they move, can effectively recognize Activities of Daily Living (ADL) in real-world rehabilitation settings. We evaluated our models on a complex dataset collected in the wild comprising 2261 minutes of egocentric video from 16 participants with impaired hand function. By leveraging pre-trained object detection and hand-object interaction models, our system achieves robust performance across different impairment levels and environments, with our best model achieving a mean weighted F1-score of 0.78 ± 0.12 and maintaining an F1-score > 0.5 for all participants using leave-one-subject-out cross validation. Through qualitative analysis, we observe that this approach generates clinically interpretable information about functional object use while being robust to patient-specific movement variations, making it particularly suitable for rehabilitation contexts with prevalent upper limb impairment.

## Introduction

Regaining hand function for Activities of Daily Living (ADLs) is a top priority for individuals with stroke or spinal cord injury (SCI) during community reintegration [1,2]. While these activities are critical for autonomy and societal integration, current clinical assessment methods rely heavily on direct observation and patient self-reporting. These traditional approaches are limited by recall bias and provide only snapshots of function in clinical settings, failing to capture the diversity of real-world environments and compensatory strategies patients develop at home [3]. This gap between clinical assessment and actual home function presents a significant challenge in designing targeted rehabilitation interventions.

Wearable egocentric cameras combined with computer vision offer a promising solution for capturing real-world hand function. We previously developed a framework [4] that monitors hand use at home by analyzing movement patterns and hand-object interactions, delivering these metrics via a clinical dashboard [5]. However, clinician feedback highlighted a critical need: while quantitative metrics are valuable, therapists require contextual information about activities and object interactions to effectively guide rehabilitation. This aligns with clinical assessment practices like the Graded Redefined Assessment of Strength Sensibility and Prehension (GRASSP) [6] or Action Research Arm Test (ARAT) [7], where therapists evaluate functional recovery through a patient's ability to manipulate objects in daily tasks.

Recent approaches in egocentric activity recognition that directly process video sequences, such as 3D Convolutional Neural Networks (CNNs) [8,9] and vision transformers [10,11], have been driven by the availability of extensive egocentric datasets like EPIC-KITCHENS [12] and Ego4D [13]. However, these approaches do not account for some key challenges in rehabilitation contexts: (1) they typically require extensive training data with consistent movement patterns, which may not be available or appropriate for patients who develop individualized compensatory strategies [14], (2) they often operate as black boxes, making their predictions difficult for therapists to interpret and trust, and (3) they are generally constrained to recognizing activities from a predetermined set, limiting their ability to capture the diverse and evolving ways patients accomplish tasks during recovery.

---

[*] Corresponding author. Address: The KITE Research Institute, University Health Network, 550 University Avenue, #12-102, Toronto, ON, M5G 2A2 Canada. Email: jose.zariffa@utoronto.ca.

To address these challenges, we propose an object-centric approach that focuses on what objects patients interact with rather than how they move—a crucial aspect of activity recognition that has been previously validated in literature [15–19]. This strategy offers three key advantages for rehabilitation settings: (1) flexibility in recognizing activity categories based on object interaction patterns rather than requiring specific predefined activities, (2) interpretable results that align with clinical assessment methods by providing clear information about functional object use, and (3) feasible deployment without requiring patient-specific training data through the use of pre-trained object detection models.

The primary contribution of our work is demonstrating that a simplified object-based approach can achieve robust activity recognition in real-world home settings across different impairment levels. This functional context supplements information about quantity [4] and quality [20,21] of hand use in existing frameworks [5]. This work bridges the gap between automated monitoring and clinical utility, providing therapists with the contextual information they need for rehabilitation planning.

# Methods

## Dataset & Preprocessing

We performed a retrospective analysis of the dataset previously described by Bandini *et al.* [4], where 21 participants recorded themselves performing an array of real ADLs within their home environments without any imposed constraints, following the recording protocol outlined by Tsai *et al.* [22].

The recordings were segmented into one-minute snippets. Video snippets were excluded from analysis if they met any of the following criteria: (1) contained sensitive or identifying information (e.g., toileting activities, faces, passwords, bank statements, etc.), (2) insufficient visibility (e.g., extremely low lighting conditions / black screen), or (3) no discernible hand or object movement (e.g., static scenes where the participant was stationary). These exclusion criteria were established to ensure data quality and reliable model performance, as segments without visible objects or interactions provide no informative features for our object-centric approach. The final dataset, after exclusions, is comprised of 2261 one minute egocentric video snippets obtained from 16 participants with impaired hand functionality (American Spinal Injury Association Impairment Scale A-D; Level of Injury C3-C7) due to SCI. The number of video snippets varied per participant, ranging from a minimum of 7 minutes to a maximum of 229 minutes, with an average duration of 141.31 ± 72.91 minutes.

Video snippets were manually classified into seven predefined ADL categories aligning with the American Occupational Therapy Association's Occupational Therapy Practice Framework [23] and chosen based on the most common activities observed in the dataset:

1. Self-Feeding (257 instances): Activities related to setting up and consuming meals, including manipulating utensils, drinking vessels, and food items. This encompasses tasks from opening containers to bringing food or drink to the mouth.
2. Functional Mobility (207 instances): Movement between positions and locations, including wheelchair mobility, transfers (e.g., bed to chair), and navigation of the home environment. This category focuses on how participants interact with mobility aids and environmental features.

3. Grooming & Health Management (172 instances): Personal care activities including hygiene routines, medication management, and exercise. This encompasses tasks such as brushing teeth, applying personal care products, and managing health-related equipment.
4. Communication Management (428 instances): Activities involving the use of communication devices and tools, such as phones, computers, or tablets. This includes tasks like typing, holding devices, and manipulating communication equipment.
5. Home Management (407 instances): Tasks related to maintaining personal and household possessions and environments. This includes cleaning, organizing, and basic home maintenance activities.
6. Meal Preparation and Cleanup (625 instances): Activities involved in planning, preparing, and serving meals, as well as cleaning up afterward. This encompasses using kitchen tools, appliances, and cleaning supplies.
7. Leisure & Other Activities (165 instances): Non-obligatory activities performed during discretionary time, including hobbies, entertainment, and social activities that involve object manipulation.

In cases where multiple ADLs were observed in a single snippet, the snippet was assigned the label of the predominant ADL (i.e., the one performed for the longest duration within the minute).

## Feature Engineering Pipeline

Our feature engineering pipeline (Figure 1) consists of three main stages: object detection, interaction detection, and feature generation.

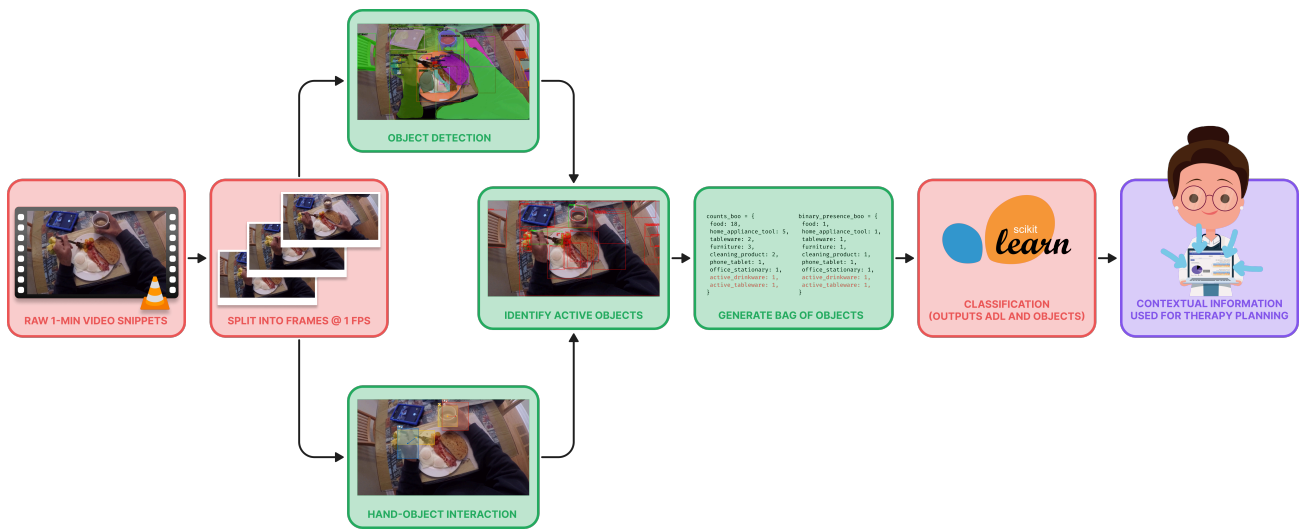

Figure 1: Pipeline for detecting ADLs from egocentric videos. Steps in red indicate operations on the entire 1-minute video snippet while steps in green indicate operations on individual frames.

### Object Detection

For object detection, we employed the pre-trained Detic model [24], which detects a broad range of objects in each video frame. The detected objects were mapped to 29 functional categories relevant to rehabilitation (e.g., "kitchen_utensils", "electronics", "wheelchair_walker") through a predefined mapping scheme to standardize object classifications.

### Interaction Detection

To identify object interactions, we utilized the pre-trained 100DOH hand-object interaction model [25]. Objects were classified as either active or passive based on their spatial relationship with detected hand interactions. Specifically, an object was considered active if a Detic bounding box had an intersection over union greater than 0.8 with a 100DOH object bounding box. This distinction is clinically relevant as it differentiates between objects patients can effectively manipulate versus those that appear passively in the scene.

**Expected Model Performance**

Both Detic [24] and 100DOH [25] were evaluated on our dataset to assess their expected performance. We used a stratified sampling approach, randomly selecting two videos from each participant for each ADL category for annotation. The resulting subset used for evaluating object detection comprised 1482 images, containing 4757 object bounding boxes, which represented approximately 5% of the total dataset [26]. Detic achieved a mean average precision (mAP) of 0.19 on all objects (i.e., both active and passive objects) and 0.30 on only active objects [26]. However, our ground truth labeling approach, which involved marking bounding boxes only on objects relevant to the ADLs rather than all objects, led to a high rate of false positives, suggesting that the actual performance of the model might be even better. The 100DOH model [25] was evaluated on 632,180 manually annotated frames from 13 participants and achieved a median F1-score of 0.80 (0.67-0.85), indicating good performance on our dataset [4].

**Feature Generation**

For each one-minute video segment, frames were extracted at 1 FPS and processed through this pipeline. From these processed frames, we generated two types of feature vectors, or Bags of Objects (BoO): (1) Binary presence vectors indicating whether each object category appeared in the frame, summed across frames in each segment; and (2) Count vectors containing the frequency of each object category's appearance in each frame, summed across frames in each segment.

Both representations underwent row-wise min-max scaling to normalize features within each segment, emphasizing the relative presence of objects rather than absolute counts. This normalization strategy makes our approach more robust to variations in recording duration and activity speed across participants with different impairment levels.

## ADL Classification

We evaluated five classification models: logistic regression (LR), random forest (RF), gradient boosting (GB), XGBoost (XGB), and multi-layer perceptron (MLP). The classification models were implemented using scikit-learn, with balanced class weights for logistic regression and random forest to handle class imbalance. The multi-layer perceptron used adaptive learning rates and early stopping to prevent overfitting. For gradient boosting and XGBoost, we used default parameters.

## Evaluation Framework

To assess generalization, we used leave-one-subject-out cross-validation [27]. This evaluation strategy is particularly important in rehabilitation contexts, where individual variations in movement patterns can significantly impact activity recognition.

Model performance was evaluated using weighted F1-score to account for class imbalance, along with the percentage of participants achieving F1-score greater than 0.5. These metrics were chosen to reflect both overall system accuracy and clinical utility—a rehabilitation monitoring system must perform consistently across different patients to be practically useful.

We conducted an ablation study comparing different feature combinations (i.e., binary presence vs. counts vs. both, with and without active object distinction) to determine the most robust approach for rehabilitation settings. Classification without the active objects were done by using the Detic [24] model's detections prior to determining active objects using the 100DOH [25] model. This analysis helps understand which aspects of object interaction are most informative for activity recognition in patients with impaired hand function, and how different feature representations impact the system's ability to handle compensatory movements.

# Results

The impact of including active object information was consistent across different feature representations (Table 1). Using logistic regression, our best performing classifier, as an example, we observe the weighted F1-score improved from 0.70 ± 0.14 to 0.73 ± 0.13 with the inclusion of active objects when using counts, from 0.73 ± 0.15 to 0.78 ± 0.12 when using binary presence, and from 0.72 ± 0.13 to 0.77 ± 0.13 when using both.

Table 1: Performance of classifiers across different feature representations.

| Feature Vector | Active Objects | Mean Weighted F1-score | % of Participants > 0.5 F1-score |
|---|---|---|---|
| Counts |  | GB: 0.66 ± 0.23<br>LR: 0.70 ± 0.14<br>MLP: 0.65 ± 0.20<br>RF: 0.65 ± 0.22<br>XGB: 0.67 ± 0.22 | GB: 81%<br>LR: 88%<br>MLP: 81%<br>RF: 75%<br>XGB: 81% |
| Counts | ✓ | GB: 0.69 ± 0.20<br>LR: 0.73 ± 0.13<br>MLP: 0.70 ± 0.18<br>RF: 0.68 ± 0.23<br>XGB: 0.70 ± 0.20 | GB: 88%<br>LR: 94%<br>MLP: 88%<br>RF: 81%<br>XGB: 88% |
| Binary |  | GB: 0.68 ± 0.18<br>LR: 0.73 ± 0.15<br>MLP: 0.65 ± 0.24<br>RF: 0.64 ± 0.24<br>XGB: 0.69 ± 0.18 | GB: 81%<br>LR: 94%<br>MLP: 81%<br>RF: 81%<br>XGB: 88% |
| Binary | ✓ | GB: 0.72 ± 0.19<br>**LR: 0.78 ± 0.12**<br>MLP: 0.73 ± 0.21<br>RF: 0.69 ± 0.24<br>XGB: 0.74 ± 0.17 | GB: 88%<br>**LR: 100%**<br>MLP: 81%<br>RF: 81%<br>XGB: 88% |
| Counts + Binary |  | GB: 0.69 ± 0.20<br>LR: 0.72 ± 0.13<br>MLP: 0.68 ± 0.18<br>RF: 0.69 ± 0.23<br>XGB: 0.68 ± 0.17 | GB: 81%<br>LR: 94%<br>MLP: 88%<br>RF: 81%<br>XGB: 81% |
| Counts + Binary | ✓ | GB: 0.71 ± 0.19<br>LR: 0.77 ± 0.13<br>MLP: 0.72 ± 0.16<br>RF: 0.69 ± 0.23<br>XGB: 0.70 ± 0.17 | GB: 81%<br>LR: 94%<br>MLP: 94%<br>RF: 81%<br>XGB: 88% |

We observed that binary presence features generally outperformed object counts across all classifiers. This suggests that for ADL classification in rehabilitation settings, the simple presence or absence of objects is more informative than their frequency. Most notably, the combination of binary presence features *with* active object distinction achieved our highest performance (F1: 0.78 ± 0.12) while maintaining F1-scores above 0.5 for all participants. This suggests that distinguishing between objects that patients can actively manipulate versus those that are merely present in their environment provides valuable information for ADL classification. This result also indicates the robustness of our method on data from new participants unseen during training.

Looking at the confusion matrix (Figure 2) from our best performing model, we observe strong performance on ADLs with distinctive object patterns. "Communication Management" shows good diagonal accuracy (0.68), with most misclassifications occurring with "Leisure & Other Activities", likely due to shared electronic device usage patterns. Similarly, "Meal Preparation and Cleanup" and "Self Feeding" show strong diagonal values (0.84 and 0.78 respectively), characterized by distinctive kitchen objects and tableware. However, the model showed lower performance on "Grooming & Health Management" and "Leisure & Other Activities", which had the fewest training samples (172 and 165 instances respectively) and more variable object patterns.

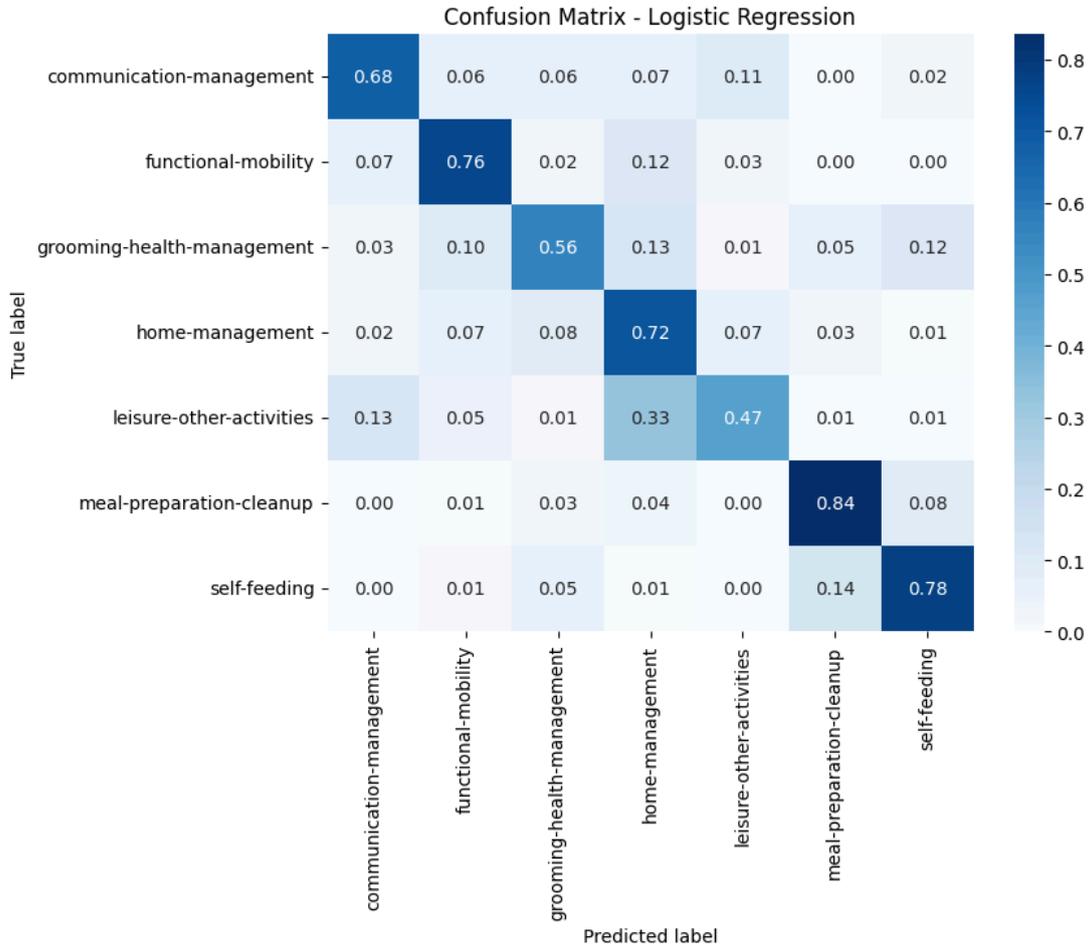

Figure 2: Confusion matrix for LR using binary presence feature vector with active objects.

We evaluated our approach's robustness to imperfect object detection by comparing performance on a subset of data (approximately 5% of our dataset) where we had manual ground truth object annotations (Table 2). Despite only modest object detection performance on our dataset, the ADL classification performance remained stable whether using automatically detected objects (F1: 0.65 ± 0.16) or ground truth annotations (F1: 0.62 ± 0.17). This suggests that our activity recognition approach is robust to object detection errors, as even imperfect object detection provides sufficient information for reliable ADL classification.

Table 2: Performance of LR trained using the binary presence feature vector with active objects on ground truth bounding boxes versus detected objects on a subset (~5%) of the data.

| Training Data | Mean Weighted F1-score | % of Participants > 0.5 F1-score |
|---|---|---|
| Ground Truth | 0.62 ± 0.17 | 69% |
| Object Detections | 0.65 ± 0.16 | 75% |

Qualitative analysis reveals cases where predicted labels, while technically incorrect, may be clinically interpretable. For example, in Figure 3 (Left), the participant is taking medication in the kitchen with drinkware present. While ground truth labels this as "Grooming & Health Management", the model's prediction of "Self Feeding" reflects the similar object patterns between medication consumption and eating activities. Such cases highlight the overlap in object usage patterns across different ADLs and suggest potential refinements in how activities are categorized for rehabilitation assessment.

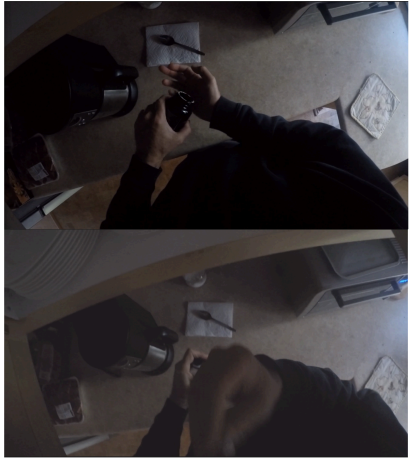
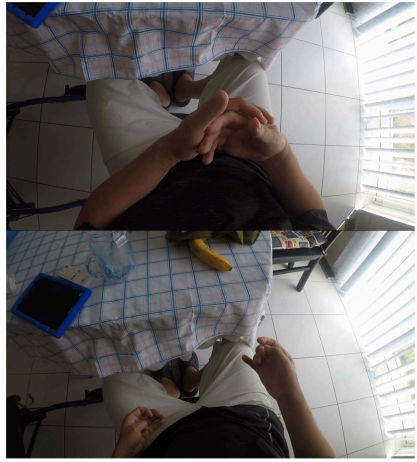
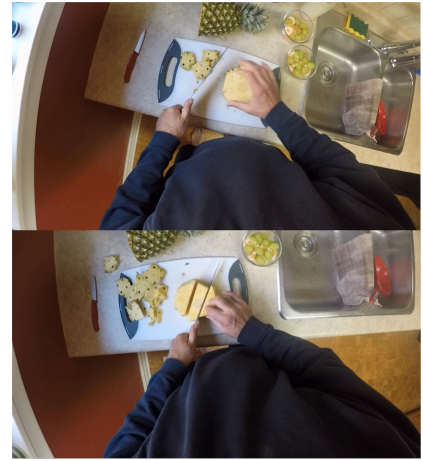

**Prediction:**
*Self-Feeding*

**Ground Truth:**
*Grooming & Health Management*

**Detected Objects:**
'active_drinkware': 0.1538
'active_phone_tablet': 0.0769
'active_electronics': 0.0769
'phone_tablet': 0.1538
'other': 0.2307
'office_stationary': 0.0769
'furniture': 0.0769
'furnishing': 0.3076
'drinkware': 0.6923
'home_appliance_tool': 0.5384
'cleaning_product': 1.0,
'bag': 0.0769
'electronics': 0.0769
'kitchen_appliance': 0.1538
'house_fixtures': 0.3846
'tableware': 1.0

**Prediction:**
*Home Management*

**Ground Truth:**
*Grooming & Health Management*

**Detected Objects:**
'clothing_accessory': 0.0769
'phone_tablet': 0.1538
'other': 0.9230
'office_stationary': 0.0769
'footwear': 0.6923
'furniture': 0.0769
'furnishing': 0.0769
'drinkware': 0.0769
'home_appliance_tool': 1.0
'food': 0.0769
'clothing': 0.3076
'cleaning_product': 0.0769
'electronics': 0.0769
'wheelchair_walker': 0.1538
'sports_equipment': 0.6153
'tv_computer': 0.0769
'house_fixtures': 0.2307
'tableware': 0.2307

**Prediction:**
*Meal Preparation & Cleanup*

**Ground Truth:**
*Meal Preparation & Cleanup*

**Detected Objects:**
'active_food': 0.7692
'clothing_accessory': 0.0769
'other': 0.1538
'furnishing': 0.0769
'drinkware': 0.3846
'home_appliance_tool': 0.5384
'food': 1.0
'cleaning_product': 1.0
'toiletries': 0.0769
'kitchen_utensils': 1.0
'sink': 1.0
'house_fixtures': 0.8461
'tableware': 1.0
'bathroom_fixture': 0.6923

Figure 3: Qualitative evaluation of ADL classification displaying sample frames from the video snippets, followed by the ADL prediction, ADL ground truth, and the min-max normalization of detected objects from our best performing model, LR with active objects and binary presence.

# Discussion

Understanding how patients use their hands in functional tasks at home is crucial for effective rehabilitation planning. While detailed analysis of movement patterns and quality remains essential, contextualizing these movements within daily activities provides therapists with crucial insights for treatment planning. Our work demonstrates that automated ADL recognition can provide this contextual layer, complementing existing approaches to movement analysis in rehabilitation settings.

Building on earlier work showing the importance of objects in activity recognition [28–31], we explored an object-centric approach to address this need for activity context. Traditional approaches to egocentric activity recognition like CNNs [32], transformers [10,11,33,34] multimodal methods [35,36], or advanced domain adaptation techniques [37], while powerful for general activity recognition, address fundamentally different objectives than our work. These methods typically predict specific action-object combinations (e.g., "cutting tomato," "opening drawer") from a predetermined list of actions. However, our goal is to provide broader activity categories that help therapists interpret the hand movement metrics they already receive through our clinical dashboard [5]. For example, knowing that a 60% interaction rate occurred during "Meal Preparation" versus "Communication Management" provides crucial context for understanding patient function, as these activities have inherently different interaction demands. Our object-centric approach aligns with this clinical need by identifying patterns of functional object use within ADL categories, which represent collections of related activities rather than specific actions. This categorization allows therapists to meaningfully compare hand function metrics across similar activities and better understand how patients engage with different types of daily tasks. Additionally, while CNNs and transformers operate as black boxes, our method provides therapists with transparent, interpretable information about which objects patients interact with during different categories of activities—information they can use to normalize quantitative metrics and identify specific videos for detailed movement quality assessment.

Our findings reveal that binary presence features consistently outperform object counts, suggesting that for ADL classification in rehabilitation contexts, the ability to interact with specific objects is more informative than interaction frequency. This may be due to inaccuracies in object detections, which may have added additional noise to the model trained on object counts, causing them to overfit to the counts of particular objects. As a result, models trained using binary presence may be robust to inaccurate object detections at the frame-level since the correct objects are detected at some point in the frames of a video snippet.

The addition of active object detection significantly improved classification performance, achieving a mean weighted F1-score of 0.78 ± 0.12 with 100% of participants maintaining scores above 0.5 with our best model. This robustness across participants is particularly important in rehabilitation, where individual variations in movement patterns and compensatory strategies can significantly impact activity recognition. The stability of our approach even with imperfect object detection suggests its viability for real-world deployment where perfect object recognition cannot be guaranteed.

However, several limitations warrant discussion. While our current approach excluded video segments with poor visibility or lack of activity during the analysis phase, a deployed system would need to handle these scenarios automatically. Future work should focus on developing robust quality control mechanisms that can automatically identify and flag unsuitable segments. This could include implementing low-light enhancement techniques [38,39] or motion thresholds [40], to automatically determine which segments should be forwarded for ADL classification. Such an automated pipeline would be essential for real-world deployment where clinicians need reliable analysis of meaningful activities while being aware of periods that couldn't be classified due to technical limitations.

Additionally, the lower performance on "Grooming & Health Management" and "Leisure Activities" highlights another challenge: activities with diverse object interaction patterns or limited training data are harder to classify. Future work could address this through targeted data collection for underrepresented activities or by refining activity categories based on object interaction patterns rather than traditional ADL definitions.

Our qualitative analysis revealed cases where activities with similar object patterns (e.g., taking medication versus eating) were confused by the classifier. While this suggests potential limitations in using objects alone for activity recognition, it also indicates that our approach captures meaningful patterns in how objects are used in different contexts. Our system's decisions can be directly traced to the presence and manipulation of specific objects. For example, a high ratio of active kitchen utensils and tableware strongly indicates meal preparation activities, while active electronics suggest communication management. This transparency is particularly valuable in clinical settings, where therapists need to understand and trust system outputs to incorporate them into treatment decisions. This interpretability also enables easier error analysis—when misclassifications occur, we can identify which object patterns led to the confusion and potentially refine our feature engineering or class definitions accordingly.

Our simplified strategy for ADL recognition—using only binary object presence and active object detection—can effectively provide the activity context needed by clinicians to interpret hand function metrics. This has practical implications for deployment in rehabilitation settings, as the direct mapping between objects and activities makes the system's decisions transparent and interpretable, which is crucial for our specific use case of contextualizing hand use metrics in our clinical dashboard [5]. Future work should investigate whether incorporating additional feature representations (e.g., room context, temporal patterns) could enhance ADL classification while maintaining this interpretability and robustness to individual variations in movement patterns. Additionally, exploring how to better integrate activity context with detailed movement analysis could provide even richer insights for rehabilitation planning.

# Conclusion

Using an object-centric approach, we demonstrate the use of object detection as a proxy for accurately identifying ADLs in egocentric video snippets of 1-min in length. Our findings reveal that the binary presence feature representation consistently outperforms object counts, suggesting that in rehabilitation settings, the ability to interact with specific objects is more informative than interaction frequency. The inclusion of active object detection, which distinguishes between objects patients can manipulate versus those that are merely present, further improves performance, achieving a mean weighted F1-score of 0.78 ± 0.12 with robust performance across all participants.

Our approach offers three key advantages for rehabilitation monitoring: (1) a classification strategy that identifies ADL categories without requiring predetermined list of action-object pairs, (2) interpretable results that show clinicians which objects patients interact with during different activities, enabling more meaningful comparison of hand function metrics, and (3) reliable ADL categorization through the use of pre-trained object detection and hand-object interaction models that is robust to imperfect classifications. While challenges remain in automating quality control and handling activities with diverse object patterns, our work demonstrates that ADL recognition based on object interactions can effectively supplement existing hand function analysis. These results suggest that providing clinicians with detailed contextual information about when and how patients use objects at home is feasible, offering new opportunities for understanding functional recovery and personalizing rehabilitation strategies based on patients' actual daily activities.